# A generalised editor calculus (Short Paper)

A. T. Mortensen
Aalborg University
Denmark
atmo20'at'student.aau.dk

B. Bennetzen
Aalborg University
Denmark
bbenne20'at'student.aau.dk

H. Hüttel*
Aalborg University
Denmark
hans'at'cs.aau.dk

N. R. Kristensen
Aalborg University
Denmark
nrkr20'at'student.aau.dk

P. B. Steffensen
Aalborg University
Denmark
psteff19'at'student.aau.dk

**Abstract**

In this paper, we present a generalization of a syntax-directed editor calculus, which can be used to instantiate a specialized syntax-directed editor for any language, given by some abstract syntax. The editor calculus guarantees the absence of syntactical errors while allowing incomplete programs. The generalized editor calculus is then encoded into a simply typed lambda calculus, extended with pairs, booleans, pattern matching and fixed points.

## 1 Introduction

This paper is on the topic of syntax-directed editors. Most current examples of syntax-directed editor calculi, such as the ones presented by [5] and [1], are typically tailored specifically towards editing abstract syntax trees from an applied $\lambda$-calculus. This fact alone limits the usable scope of these calculi, since they would be rendered useless given any other form of abstract syntax.

Furthermore, these calculi do not account for binding mechanisms in abstract syntax. Many of the editor calculi possess the ability to interleave construction and evaluation of programs, however, a consequence of this is a complex type system.

In this paper we propose a generalized editor calculus that can be used to create a syntax-directed editor calculus for any abstract syntax. Our editor calculus is inspired by higher-order abstract syntax presented in [6], and it will take binding-mechanisms into account. We have encoded our generalized editor calculus into a simply-typed lambda calculus extended with pairs, booleans, pattern matching and fixed-points.

## 2 Generalized Editor Calculus

We will now present a generalization of the syntax-directed editor calculus by [1]. This specific editor calculus was designed with the language of an applied $\lambda$-calculus in mind.

### 2.1 Syntax

**2.1.1 Abstract Syntax.** Firstly, we introduce the general representation of some abstract syntax that will be the focus of the editor calculus. We assume the abstract syntax is given by a set of sorts $\mathcal{S}$, an arity-indexed family of operators $\mathcal{O}$ and a sort-indexed family of variables $\mathcal{X}$, such as presented in [2]. For the full definitions we refer to the full version of the paper.

The notion of cursors and holes is central to the idea of an editor calculus. Whereas the calculus in [1] only has one cursor and hole term, in the general case we need a cursor and hole operator for every sort in the abstract syntax. With the set of sorts $\mathcal{S}$, the extended family of operators $\mathcal{O}$ and family of variables $\mathcal{X}$ we can then define the sort-indexed family of abstract binding trees (refered to as "abt") $\mathcal{BX}$ as the smallest family that satisfies the conditions of $\mathcal{BX}$'s in [2].

---
*Supervisor

**Example 2.1.** To illustrate how the generalized editor calculus could be instantiated for some language, we will through our examples specialize an editor calculus. The abstract syntax for our specialization language is shown in fig. 1.

| Sort | | Term | Operator | Arity |
|---|---|---|---|---|
| $s$ | ::= | let $x = e$ in $s$ | let | $e, e.ss$ |
| | \| | $e$ | exp | $es$ |
| $e$ | ::= | $e_1\ e_2$ | plus | $e, ee$ |
| | \| | $n$ | num$n$ | $e$ |
| | \| | $x$ | var$x$ | $e$ |

**Figure 1.** Abstract syntax of arithmetic expressions and local declarations

The syntax in fig. 1 describes a very simple language of statements and expressions. For example, using the abstract syntax we could construct the statement

$$\text{let } x = 5 \text{ in let } y = 10 \text{ in } x\ y$$

which, using operators, would be represented as the abt

$$\text{let}5; x.\text{let}10; y.\text{exp}\text{plus}x; y$$

Given the abstract syntax in fig. 1 we can extend it with holes and cursors. This gives us the new operators shown in fig. 2.

| Sort | | Term | Operator | Arity |
|---|---|---|---|---|
| $s$ | ::= | $s$ | cursor$_s$ | $ss$ |
| | \| | $(\!(\ )\!)_s$ | hole$_s$ | $s$ |
| $e$ | ::= | $e$ | cursor$_e$ | $ee$ |
| | \| | $(\!(\ )\!)_e$ | hole$_e$ | $e$ |

**Figure 2.** Introduction of cursors and holes to the abstract syntax in fig. 1

With this extension we can now represent statements with holes and cursors as abt's. For example the statement

$$\text{let } x = 5 \text{ in } x\ (\!(\ )\!)_e$$

would be represented as the abt

$$\text{let}\text{cursor}_e 5; x.\text{exp}\text{plus}x; \text{hole}_e$$

**2.1.2 Editor Calculus.** The abstract syntax of the generalized editor calculus is shown in fig. 3. It closely resembles that of the calculus presented by [1].

$$
\begin{aligned}
E &::= \pi.E \mid \phi \Rightarrow E_1|E_2 \mid E_1 \ggg E_2 \mid \text{rec } x.E \mid x \mid \text{nil} \\
\pi &::= \text{child } n \mid \text{parent} \mid \{o\} \\
\phi &::= \neg\phi \mid \phi_1 \wedge \phi_2 \mid \phi_1 \vee \phi_2 \mid @o \mid \Diamond o \mid \Box o
\end{aligned}
$$

**Figure 3.** Abstract syntax of general editor calculus

Editor expressions $E \in Edt$, where $Edt$ is the sort of editor expressions, describes the behaviour of the editor calculus. The semantics behind the different constructs are defined in section section 2.2.1.

Evaluation of abt's, which [1] introduces through a *eval* construct, is not included in our generalized version. This omission is due to evaluation being entirely dependent on the dynamics of the language in focus, and not the static and structural properties the abstract syntax describes. To define the concept of evaluation in a generalized manner, it would be necessary to devise a formal and concise approach for specifying the semantics of operators.

**Example 2.2.** Expanding upon example 2.1, we can specialize the abstract syntax of the general editor calculus presented in fig. 3 for our simple language by defining the set of operators $o$ ranges over. In this case we have that

$$o \in \{\text{let}, \text{exp}, \text{hole}_s, \text{plus}, \text{num}n, \text{var}x, \text{hole}_e\}$$

allowing us to write editor expressions such as @hole$_e \Rightarrow \{\text{plus}\}.\text{nil}|\text{nil}$, which would substitute the current tree encapsulated by the cursor with the *plus* operator, if the cursor is at the operator hole$_e$.

**2.1.3 Cursor Context.** In editor expressions, we frequently refer to performing actions on, or with, the abt that is currently encapsulated by the cursor. To support this notion we introduce cursor contexts $C$, inspired by the zipper data structure by [3], as a way to locate the cursor in an abt. Central to this idea is the cursor context $\cdot$, or context hole, which specifies the subtree within which the cursor must reside, either as the root, or one of the first children. For the full definition of cursor contexts, we refer to the full version of the paper, or to [1], as they have a very similar construct.

Although a cursor context accurately specifies the subtree where the cursor must reside, it does not inherently guarantee the presence of only one cursor within that subtree. Therefore, it becomes necessary to introduce the concept of well-formed abt's. We will do this informally, where a well-formed abt is simply an abt containing exactly one cursor. This is guaranteed when an abt can be interpreted as $Ca$, meaning the context C, where $\cdot$ is substituted with a tree $a$, where the cursor-operator is either the root $a$, or one of the children of the root.

**Example 2.3.** We will now show the extension of fig. 1, where cursors are included, as shown in fig. 4. We note that $\hat{s}$ and $\hat{e}$ are trees without cursors.

| $s$ | ::= | let $x = \hat{e}$ in $\hat{s}$ \| let $x = \hat{e}$ in $\hat{s}$ \| $\hat{e}$ \| $\hat{s}$ |
|---|---|---|
| $e$ | ::= | $\hat{e}_1\ \hat{e}_2$ \| $\hat{e}_1\ \hat{e}_2$ \| $\hat{e}$ |

**Figure 4.** Abstract syntax of abt's with exactly one cursor

Given the syntax in fig. 4 and the previously introduced cursor context, we can now determine the well-formedness of any abt in our language. For example the statement

$$\text{let } x = (\!|\ |\!)_e \text{ in } x\ 5$$

is well-formed since it can be written as $Ce$ where

$$C = \text{let } x = \cdot \text{ in } x\ 5$$
$$e = (\!|\ |\!)_e$$

It should be noted that there often are two valid interpretations of a tree $a$ as $Ca$. In this example, the other valid interpretation would be:

$$C' = \cdot$$
$$e' = \text{let } x = (\!|\ |\!)_e \text{ in } x\ 5$$

### 2.2 Semantics

In this section we present the transition systems and the general forms of transition rules defining these systems, based on the generalized syntax of the editor calculus previously presented.

#### 2.2.1 Editor Expressions.
The labelled transition system for editor expressions is defined as $Edt \times \mathcal{BX}, Apc \cup \{\epsilon\}, \Rightarrow$. Transitions are of the form $\langle E, a \rangle \stackrel{\alpha}{\Rightarrow} \langle E', a' \rangle$ where the editor expression $E \in Edt$ is closed and the abt $a \in \mathcal{BX}$ is well-formed. The labels of the transitions, $\alpha$, are the atomic prefix commands $Apc$ and the silent transition $\epsilon$. The transition relation $\Rightarrow$ is defined by the rules shown in fig. 5. These facilitate ways of composing editor expressions together with conditions, sequential composition and recursion. The means of traversing and modifying an abt are provided by the prefixed expression $\pi.E$, which evaluates $\pi$ before continuing with $E$. The conditional expression $\phi \Rightarrow E_1 | E_2$ reduces to $E_1$ if $\phi$ is satisfied and $E_2$ otherwise. The sequential expression $E_1 \ggg E_2$ evaluates $E_2$ only once $E_1$ has been reduced to $nil$. The recursive expression $rec\ x.E$ binds the recursion variable $x$ in $E$, which can then be used to recursively iterate an expression.

Unlike the editor calculus by [1], we do not utilize any form of structural congruence on editor expressions to define the semantics of trivial sequential composition or recursion. Instead we simply define these notions with their own transition rule (Seq-Trivial) and (Recursion)

$$\text{(Cond-1)}\ \frac{a \models \phi}{\langle \phi \Rightarrow E_1 | E_2, Ca \rangle \stackrel{\epsilon}{\Rightarrow} \langle E_1, Ca \rangle}$$

$$\text{(Cond-2)}\ \frac{a \not\models \phi}{\langle \phi \Rightarrow E_1 | E_2, Ca \rangle \stackrel{\epsilon}{\Rightarrow} \langle E_2, Ca \rangle}$$

$$\text{(Seq)}\ \frac{\langle E_1, a \rangle \stackrel{\alpha}{\Rightarrow} \langle E_1', a' \rangle}{\langle E_1 \ggg E_2, a \rangle \stackrel{\alpha}{\Rightarrow} \langle E_1' \ggg E_2, a' \rangle}$$

$$\text{(Seq-Trivial)}\ \frac{}{\langle \text{nil} \ggg E_2, a \rangle \stackrel{\epsilon}{\Rightarrow} \langle E_2, a \rangle}$$

$$\text{(Recursion)}\ \frac{}{\langle rec\ x.E, a \rangle \stackrel{\epsilon}{\Rightarrow} \langle Ex := rec\ x.E, a \rangle}$$

$$\text{(Context)}\ \frac{a \stackrel{\pi}{\Rightarrow} a'}{\langle \pi.E, Ca \rangle \stackrel{\pi}{\Rightarrow} \langle E, Ca' \rangle}$$

**Figure 5.** Reduction rules for editor expressions

#### 2.2.2 Substitution and cursor movement.
The labelled transition system for substitutions and cursor movement is defined as $\mathcal{BX}, Apc, \Rightarrow$. This system describes the semantics of modifying the well-formed abt $a$ encapsulated by the cursor, by either substituting it with an operator or moving the cursor up or down the tree. Transitions are therefore of the form $a \stackrel{\pi}{\Rightarrow} a'$, where $a, a' \in \mathcal{BX}$ and $\pi \in Apc$.

For substitution we define the transition rule for every label $\{o\} \in Apc$, which substitutes the abt currently encapusulated by the cursor with $o$. The general case of this rule is shown in fig. 6. The side-condition ensures that we can only substitute operators of sort $s$ with abt's of sort $s$.

$$\text{(Ins)}\ \frac{}{\hat{a} \stackrel{\{o\}}{\Rightarrow} o\overrightarrow{x_1}.(\!|\ |\!)_{s_1}; \ldots; \overrightarrow{x_n}.(\!|\ |\!)_{s_n}}\ \hat{a}, o \in \mathcal{BX}_s$$

**Figure 6.** General form of reduction rule for substitution

For cursor movement we define two transition rules "child $i$" and "parent" for every operator $o$ of arity $\overrightarrow{s_1}.s_1, \ldots, \overrightarrow{s_n}.s_n s$ and for every $1 \le i \le n$. These rules, shown in fig. 7, facilitate cursor movement from the parent operator to child $i$ and from child $i$ back to the parent operator, respectively.

$$(\text{Child-}i) \; \dfrac{}{\hat{o}\overrightarrow{x_1}.\hat{a}_1; \ldots ; \overrightarrow{x_n}.\hat{a}_n \overset{\text{child } i}{\Rightarrow} o\overrightarrow{x_1}.\hat{a}_1; \ldots ; \overrightarrow{x_i}.a_i; \ldots ; \overrightarrow{x_n}.\hat{a}_n}$$

$$(\text{Parent}) \; \dfrac{}{o\overrightarrow{x_1}.\hat{a}_1; \ldots ; \overrightarrow{x_i}.\hat{a}_i; \ldots ; \overrightarrow{x_n}.\hat{a}_n \overset{\text{parent}}{\Rightarrow} \hat{o}\overrightarrow{x_1}.\hat{a}_1; \ldots ; \overrightarrow{x_n}.\hat{a}_n}$$

**Figure 7.** General form of reduction rules for cursor movement

**Example 2.4.** Continuing from example 2.3, we now define the semantics of the editor calculus for our simple language. We do not show the transition rules for editor expressions, since these are equivalent to the ones presented in fig. 5.

The transition system for cursor movement and substitution is defined with respect to the operators $o$ in our language. Based on the general form of substitution rules shown in fig. 6, we define a specialized rule for substituting the tree encapsulated by the cursor for every operator $o$. A selection of these are shown in fig. 8.

$$(\text{let}) \; \dfrac{}{\hat{a} \overset{\{\text{let}\}}{\Rightarrow} \text{let}(\!|\;|\!)_e; x.(\!|\;|\!)_s} \; \hat{a} \in \mathcal{B}\mathcal{X}_s$$

$$(\text{plus}) \; \dfrac{}{\hat{a} \overset{\{\text{plus}\}}{\Rightarrow} \text{plus}(\!|\;|\!)_e; (\!|\;|\!)_e} \; \hat{a} \in \mathcal{B}\mathcal{X}_e$$

$$(\text{var}) \; \dfrac{}{\hat{a} \overset{\{\text{var } x\}}{\Rightarrow} x} \; \hat{a} \in \mathcal{B}\mathcal{X}_e$$

**Figure 8.** Selected reduction rules for substitution

Notice that in every rule we ensure that the substitution can only be performed if the abt $\hat{a}$ is of the same sort as the operator. For example, given the configuration $\langle \{let\}.nil, \text{let } x = (\!|\;|\!)_e \text{ in } x \; x \rangle$ we cannot substitute in a statement, as shown below:

$$(\text{Context}) \dfrac{(\text{let}) \dfrac{}{(\!|\;|\!)_e \overset{\{\text{let}\}}{\not\Rightarrow}} \; (\!|\;|\!)_e \notin \mathcal{B}\mathcal{X}_s}{\langle \{\text{let}\}.\text{nil}, \text{let } x = (\!|\;|\!)_e \text{ in } x \; x \rangle \overset{\{\text{let}\}}{\not\Rightarrow}}$$

Similarly, using the general form of cursor movement transition rules in fig. 7, we define specialized rules for every argument $i$ of every operator $o$. In fig. 9 we show the parent and child rules for the let operator specifically. The transition rules for the remaining operators would be defined analogously to these.

$$(\text{letc-1}) \; \dfrac{}{\text{let}a_1; x.a_2 \overset{\text{child } 1}{\Rightarrow} \text{let}a_1; x.a_2}$$

$$(\text{letc-2}) \; \dfrac{}{\text{let}a_1; x.a_2 \overset{\text{child } 2}{\Rightarrow} \text{let}a_1; x.a_2}$$

$$(\text{letp-1}) \; \dfrac{}{\text{let}a_1; x.a_2 \overset{\text{parent}}{\Rightarrow} \text{let}a_1; x.a_2}$$

$$(\text{letp-2}) \; \dfrac{}{\text{let}a_1; x.a_2 \overset{\text{parent}}{\Rightarrow} \text{let}a_1; x.a_2}$$

**Figure 9.** Reduction rules for cursor movement on the *let* operator

**2.2.3 Conditions.** Finally, we define the satisfaction relation for conditions $\phi$ in our editor calculus. The propositional connectives are defined as expected in fig. 10.

$$(\text{Negation}) \; \dfrac{\hat{a} \not\models \phi}{\hat{a} \models \neg \phi}$$

$$(\text{Conjunction}) \; \dfrac{\hat{a} \models \phi_1 \quad \hat{a} \models \phi_2}{\hat{a} \models \phi_1 \wedge \phi_2}$$

$$(\text{Disjunction-1}) \; \dfrac{\hat{a} \models \phi_1}{\hat{a} \models \phi_1 \vee \phi_2}$$

$$(\text{Disjunction-2}) \; \dfrac{\hat{a} \models \phi_2}{\hat{a} \models \phi_1 \vee \phi_2}$$

**Figure 10.** Satisfaction relation for propositional connectives

The general form of the satisfaction rules for the modalities @$o$, $\Diamond o$ and $\Box o$ are defined in fig. 11.

$$(\text{At-op}) \; \dfrac{}{o\overrightarrow{x_1}.\hat{a}_1; \ldots ; \overrightarrow{x_n}.\hat{a}_n \models @o}$$

$$(\text{Necessity}) \; \dfrac{\hat{a}_1 \models \Diamond o \; \ldots \; \hat{a}_n \models \Diamond o}{o\overrightarrow{x_1}.\hat{a}_1; \ldots ; \overrightarrow{x_n}.\hat{a}_n \models \Box o}$$

$$(\text{Possibly-}i) \; \dfrac{\hat{a}_i \models \Diamond o}{o\overrightarrow{x_1}.\hat{a}_1; \ldots ; \overrightarrow{x_i}.\hat{a}_i; \ldots ; \overrightarrow{x_n}.\hat{a}_n \models \Diamond o}$$

$$(\text{Possibly-trivial}) \; \dfrac{\hat{a} \models @o}{\hat{a} \models \Diamond o}$$

**Figure 11.** Satisfaction relation for modal operators

The modal operator @$o$ is satisfied when the root of the abt encapsulated by the cursor is the operator $o$. Therefore we define the rule (At-op) for every operator $o$.

The modal operator $\Diamond o$ is satisfied when the operator $o$ is anywhere within the abt encapsulated by the cursor. For the trivial case, when the root of the tree encapsulated by the cursor is $o$, we only define the one rule (Possibly-trivial). For the non-trivial case, where $o$ is in one of the subtrees $\hat{a}_i$, we define a rule (Possibly-i) for every argument of every operator $o$.

The modal operator $\Box o$ is satisfied when $o$ is somewhere in every subtree of the abt encapsulated by the cursor. To describe this, we define the rule (Necessity) for every operator $o$.

**Example 2.5.** We finalize our specialized editor calculus from example 2.4, by defining the satisfaction relation. The propositional connectives are defined as previously shown in fig. 10. In fig. 12 we have shown the satisfaction rules for all modalities for the *plus* operator. Rules for the remaining operators $o$ are defined accordingly.

$$(\text{At-plus}) \; \frac{}{\text{plus}\hat{a}_1;\hat{a}_2 \models @\text{plus}}$$

$$(\text{Pos1-plus}) \; \frac{\hat{a}_1 \models \Diamond o}{\text{plus}\hat{a}_1;\hat{a}_2 \models \Diamond o}$$

$$(\text{Nec-plus}) \; \frac{\hat{a}_1 \models \Diamond o \quad \hat{a}_2 \models \Diamond o}{\text{plus}\hat{a}_1;\hat{a}_2 \models \Box o}$$

$$(\text{Pos2-plus}) \; \frac{\hat{a}_2 \models \Diamond o}{\text{plus}\hat{a}_1;\hat{a}_2 \models \Diamond o}$$

**Figure 12.** Satisfaction relation for modal operators on *plus*

## 3 Encoding the generalized editor calculus in an extended lambda calculus

In this section we attempt to soundly encode the generalized editor calculus using the simply typed $\lambda$-calculus ($\lambda_\rightarrow$) as the base calculus. The calculus will be extended with various constructs such as pattern matching, pairs and the recursive *fix* operator. To encode the abstract binding trees, we will only need the simply typed lambda calculus. To encode the contexts, we introduce pairs. To encode atomic prefix commands we further extend the calculus with pattern matching ($\lambda_{\rightarrow,p}$). Lastly to encode modal logic and editor expressions we extend the calculus with a fixed point operator ($\lambda_{\rightarrow,p,fix}$).

### 3.1 Motivation

The motivation behind encoding the generalized editor calculus in $\lambda_{\rightarrow,p,fix}$ (or subsets thereof) is most importantly the type system it provides. If our encoding is sound, then any instance of the editor calculus will have a sound type system, regardless of the abstract syntax. This follows from the fact that the type system of $\lambda_{\rightarrow,p,fix}$ is sound. Our editor calculus having a sound type system means that it will reject incorrect editor expressions.

Secondly a $\lambda_{\rightarrow,p,fix}$ encoding provides a clear strategy for implementing any instantiation of the generalized editor calculus in a functional programming language.

### 3.2 Abstract Binding Trees

To encode abstract binding trees we only need to use the simply typed lambda-calculus. We extend the lambda-calculus with term constants $o$ for every $o \in \mathcal{O}$ excluding cursors, and the base types $s$ for every $s \in \mathcal{S}$. Typing rules for operators can be seen on fig. 13, where we can infer the type of an operator in the lambda-calculus by its arity.

$$(\text{T-Operator}) \; \frac{o \in \mathcal{O} \text{ and has arity } \overrightarrow{s_1}.s_1,\ldots,\overrightarrow{s_n}.s_n s}{\Gamma \vdash o : \overrightarrow{s_1} \rightarrow s_1 \rightarrow \ldots \overrightarrow{s_n} \rightarrow s_n \rightarrow s}$$

**Figure 13.** Typing rules for operators.

With the operators added as term constants and sorts added as base types, encoding abts is straightforward as we just have to curry the operator $o$, encode the children $a_1, \ldots, a_n$ and add typings. This can be seen on fig. 14.

$$[\![o\overrightarrow{x_1}.a_1,\ldots,\overrightarrow{x_n}.a_n]\!] = o \; \lambda\overrightarrow{x_1}:\overrightarrow{s_1}.[\![a_1]\!] \ldots \lambda\overrightarrow{x_n}:\overrightarrow{s_n}.[\![a_n]\!]$$

**Figure 14.** Encoding of abstract binding trees.

**Example 3.1.** Continuing from example 2.5 we will now encode the operators in our small language. Following our singular rule for encoding abt's we get the resulting encoding seen on fig. 15, and the type for each operator on fig. 16.

$$[\![\text{plus}a_1;a_2]\!] = \text{plus}[\![a_1]\!][\![a_2]\!] \quad [\![\text{num}n]\!] = \text{num}n$$

$$[\![\text{var}x]\!] = \text{var}x \quad [\![\text{hole}_e]\!] = \text{hole}_e$$

$$[\![\text{exp}a_1]\!] = \text{exp}[\![a_1]\!] \quad [\![\text{hole}_s]\!] = \text{hole}_s$$

$$[\![\text{let}a_1;x.a_2]\!] = \text{let} \; [\![a_1]\!]\lambda x:e.[\![a_2]\!]$$

**Figure 15.** Encoding of operators.

$$\text{plus} : e \rightarrow e \rightarrow e \quad \text{num}n : e$$

$$\text{var}x : e \quad \text{hole}_e : e$$

$$\text{exp} : e \rightarrow s \quad \text{hole}_s : s$$

$$\text{let} : e \rightarrow e \rightarrow s \rightarrow s$$

**Figure 16.** Types for the encoded operators.

for example, if we were to encode the statement:

$$\text{let } x = (\!|\ |\!)_e \text{ in } x\ 5$$

it could be encoded as:

$$\text{let cursor}_e \text{hole}_e\ \lambda x.\text{plus var}x\ \text{num}5$$

### 3.3 Cursor Contexts

To represent abts with cursor contexts we introduce the pair construct similarly to [7].

In fig. 17 the encoding for cursor contexts can be seen. We can encode a cursor context $Ca$ as a pair of two encoded trees the first being a regular abt and the second being an abt with a context hole $\cdot$, which we encode as an operator $\cdot$. We also introduce a type alias such that $[\![Ca]\!]$ has type $Ctx = s \times s$.

$$[\![Ca]\!] = [\![a]\!], [\![C]\!]$$
$$[\![\cdot]\!] = \cdot$$

**Figure 17.** Encoding of cursor contexts.

### 3.4 Atomic Prefix Commands

To encode the atomic prefix commands we further extend the lambda calculus with pattern matching. As seen in fig. 18 we introduce the match construct where we match the term $M$ on the patterns $p$ resulting in the corresponding term $N$ if there is a match. As mentioned $p$ is the syntactic category describing the pattern we are trying to match. These patterns consist of variables, wildcards, operators, pairs, and lastly bindings which recursively match on the body of an abstraction.

|  |  | Terms |  |
|---|---|---|---|
| $M, N$ | ::= | match $M\ \overrightarrow{p \to N}$ | (match construct) |
| $p$ | ::= | $x$ | (variable) |
|  | \| | _ | (wildcard) |
|  | \| | $o\ \overrightarrow{p}$ | (operator) |
|  | \| | $p_1, p_2$ | (pair) |
|  | \| | $.p$ | (binding) |

**Figure 18.** Abstract syntax for representing atomic prefix commands, extending the lambda-calculus with a match construct.

The reduction rules for the match construct can be seen in the full version of the paper.

We now define the auxiliary functions as seen in fig. 19. These function as the operations of a zipper data structure, which helps encode atomic prefix commands on abts. We also introduce the abbreviation $a$ to mean the cursor operator with the child $a$.

$$\text{down} \stackrel{\text{def}}{=} \lambda x : s.\text{match } x \quad o\ .a_1\ldots.a_n \to$$
$$o\ \lambda\overrightarrow{x_1}.a_1\ldots\overrightarrow{x_n}.a_n$$

$$\text{right} \stackrel{\text{def}}{=} \lambda x : s.\text{match } x \quad o\ .a_1\ \ldots.a_i\ \ldots.a_n \to$$
$$o\ \lambda\overrightarrow{x_1}.a_1\ \ldots\lambda\overrightarrow{x_{i1}}.a_{i1}\ \ldots\overrightarrow{x_n}.a_n$$

$$\text{up} \stackrel{\text{def}}{=} \lambda x : s.\text{match } x \quad o\ .a_1\ldots.a_i\ldots.a_n \to$$
$$o\ \lambda\overrightarrow{x_1}.a_1\ \ldots\lambda\overrightarrow{x_n}.a_n$$

$$\text{set} \stackrel{\text{def}}{=} \lambda a : s.\lambda x : s.\text{match } x \quad a' \to a$$

**Figure 19.** Function definitions for cursor movement and substitution on abstract binding trees.

The encoding of atomic prefix commands can be seen in fig. 20. The *child* command can be encoded as a base case and a recursive case. In the base case we want to move to the first *child*, which means we just go down. In the recursive case we want to move to *child n*, we can encode this as moving to *child n* − 1 and then moving right. *parent* and *insert* can directly be encoded as *up* and *set* respectively.

$$[\![\text{child 1}]\!] = \text{down} \quad [\![\text{child } n]\!] = \text{right } [\![\text{child } n - 1]\!]$$

$$[\![\text{parent}]\!] = \text{up} \quad [\![\text{insert } a]\!] = \text{set } [\![a]\!]$$

**Figure 20.** Encoding of atomic prefix commands.

### 3.5 Control Structures and Modal Logic

To encode control structures and modal logic, we need to further extend the lambda-calculus with a fix operator as well as the *Bool* base type. We introduce the fix operator as seen in [7], as well as boolean types and values. On fig. 21 the encoding of editor expressions can be seen. For atomic prefixes the atomic prefix command should be applied before the rest of the editor expression. The *nil* editor expression will become the identity function. Sequential editor expressions will apply the encoding of the editor expressions in the correct order. Recursion works by using the newly introduced fix operator. Lastly, conditional editor expressions are encoded such that we match on the evaluation of $\phi$ on the first projection of a context $C$ and apply the encoding of $E_1$ or $E_2$ on $C$ accordingly. For the full encoding of modal logic, we refer to the full version of the paper.

$$[\![\pi.E]\!] = \lambda C : \text{Ctx}.[\![E]\!][\![\pi]\!]\ C.1\ , C.2$$

$$[\![E_1 \ggg E_2]\!] = \lambda C : \text{Ctx}.[\![E_2]\!][\![E_1]\!]C$$

$$[\![\text{rec } x.E]\!] = \text{fix } \lambda x : \text{Ctx} \to \text{Ctx}.[\![E]\!]$$

$$[\![\phi \Rightarrow E_1 | E_2]\!] = \lambda C : \text{Ctx}.\text{match } [\![\phi]\!]\ C.1$$
$$|\ \top \to [\![E_1]\!]\ C$$
$$|\ \bot \to [\![E_2]\!]\ C$$

$$[\![\langle E, Ca'\rangle]\!] = [\![E]\!]\ [\![a']\!], [\![C]\!]$$

$$[\![\text{nil}]\!] = \lambda C : \text{Ctx}.C \qquad [\![x]\!] = x$$

**Figure 21.** Encoding of editor expressions and context configuration.

With this, the entire editor calculus has been encoded. We refer to the full version of the paper to see the proof of the soundness of our encoding.

## 4 Conclusion & Further Work

We have developed a generalized editor calculus that enables the creation of a syntax-directed editor calculus for a specific abstract syntax. Subsequently, we encoded this editor calculus into an extended version of the lambda calculus, which incorporates pairs, recursion and pattern matching. Further work on our generalized editor calculus could be to prove whether the encoding is complete. Otherwise, one could implement a redo-undo construct such as introduced in [4] in the generalized editor calculus. Another interesting construct to implement would be a copy-paste operation.

## References

[1] Christian Godiksen, Thomas Herrmann, Hans Hüttel, Mikkel Korup Lauridsen, and Iman Owliaie. 2021. A Type-Safe Structure Editor Calculus. In *Proceedings of the 2021 ACM SIGPLAN Workshop on Partial Evaluation and Program Manipulation* (Virtual, Denmark) (*PEPM* 2021). Association for Computing Machinery, New York, NY, USA, 1–13. https://doi.org/10.1145/3441296.3441393

[2] Robert Harper. 2016. *Practical Foundations for Programming Languages* (2nd ed.). Cambridge University Press, USA.

[3] Gérard Huet. 1997. The Zipper. *J. Funct. Program.* 7 (09 1997), 549–554. https://doi.org/10.1017/S0956796897002864

[4] Rasmus Rendal Kjær, Magnus Holm Lundbergh, Magnus Mantzius Nielsen, and Hans Hüttel. 2021. An Editor Calculus With Undo/Redo. In *Proceedings of 23rd International Symposium on Symbolic and Numeric Algorithms for Scientific Computing (SYNASC)* (*Proceedings - 2021 23rd International Symposium on Symbolic and Numeric Algorithms for Scientific Computing, SYNASC* 2021), Carsten Schneider, Mircea Marin, Viorel Negru, and Daniela Zaharie (Eds.). IEEE, United States, 66–74. https://doi.org/10.1109/SYNASC54541.2021.00023 Publisher Copyright: © 2021 IEEE.; 23rd International Symposium on Symbolic and Numeric Algorithms for Scientific Computing (SYNASC), SYNASC ; Conference date: 07-12-2021 Through 10-12-2021.

[5] Cyrus Omar, Ian Voysey, Michael Hilton, Jonathan Aldrich, and Matthew A. Hammer. 2017. Hazelnut: A Bidirectionally Typed Structure Editor Calculus. *SIGPLAN Not.* 52, 1 (jan 2017), 86–99. https://doi.org/10.1145/3093333.3009900

[6] F. Pfenning and C. Elliott. 1988. Higher-Order Abstract Syntax. In *Proceedings of the ACM SIGPLAN 1988 Conference on Programming Language Design and Implementation* (Atlanta, Georgia, USA) (*PLDI '88*). Association for Computing Machinery, New York, NY, USA, 199–208. https://doi.org/10.1145/53990.54010

[7] Benjamin C. Pierce. 2002. *Types and Programming Languages* (1st ed.). The MIT Press.